%% file: ijcb2017.tex
\documentclass[10pt,twocolumn,letterpaper]{article}

\usepackage{btas}
\usepackage{times}
\usepackage{epsfig}
\usepackage{graphicx}
\usepackage{amsmath}
\usepackage{amssymb}
\usepackage{algorithm}
\usepackage{subfig}
\usepackage{algpseudocode}
\usepackage{subfloat}

\usepackage{booktabs}
\usepackage{multirow}
\usepackage{makecell}
\usepackage{url}

\usepackage{fancyhdr}

\usepackage{ifdraft}
\ifoptionfinal{%
\newcommand\todo[1]{}%
}{%
\newcommand\todo[1]{\textcolor{red}{\textit{TODO: #1}}}%
}

\usepackage[addedmarkup=uline]{changes}
\setremarkmarkup{\textit{(#2)}}

\renewcommand\cap[3]{\caption[#2]{\label{#1}\textsc{#2} \small{#3}}}
\graphicspath{{./images/}{./images/teaser/}{./images/adversaries/}{./images/adv_examples/}{./images/exts_on_ints/}{./images/roc/}}
\newcommand\ic[2][1]{\includegraphics[width=#1\textwidth]{#2}}

\renewcommand\etal[1]{\textit{et al.}~\cite{#1}}
\renewcommand\sec[1]{Sec.~\ref{sec:#1}}
\newcommand\fig[1]{Fig.~\ref{fig:#1}}

\newcommand\tab[1]{Tab.~\ref{tab:#1}}
\newcommand\alg[1]{Alg.~\ref{alg:#1}}

\usepackage[pagebackref=true,breaklinks=true,letterpaper=true,colorlinks,bookmarks=false]{hyperref}

\btasfinalcopy % *** Uncomment this line for the final submission

\def\btasPaperID{53} % *** Enter the IJCB Paper ID here

\ifbtasfinal\fi

\begin{document}

%%%%%%%%% TITLE
\title{LOTS about Attacking Deep Features}

\author{Andras Rozsa, Manuel G\"unther, and Terrance E. Boult \\
Vision and Security Technology (VAST) Lab \\
University of Colorado, Colorado Springs, USA\\
\small\texttt{\{arozsa,mgunther,tboult\}@vast.uccs.edu}
}

%\and

\maketitle
\thispagestyle{empty}

\chead{\footnotesize This is a pre-print of the original paper accepted at the International Joint Conference on Biometrics (IJCB) 2017.}
\lhead{}
\thispagestyle{fancy}
\pagenumbering{gobble}

%%%%%%%%% ABSTRACT

\begin{abstract}

Deep neural networks provide state-of-the-art performance on various tasks and are, therefore, widely used in real world applications.
DNNs are becoming frequently utilized in biometrics for extracting deep features, which can be used in recognition systems for enrolling and recognizing new individuals.
It was revealed that deep neural networks suffer from a fundamental problem, namely, they can unexpectedly misclassify examples formed by slightly perturbing correctly recognized inputs.
Various approaches have been developed for generating these so-called adversarial examples, but they aim at attacking end-to-end networks.
For biometrics, it is natural to ask whether systems using deep features are immune to or, at least, more resilient to attacks than end-to-end networks.
In this paper, we introduce a general technique called the layerwise origin-target synthesis (LOTS) that can be efficiently used to form adversarial examples that mimic the deep features of the target.
We analyze and compare the adversarial robustness of the end-to-end VGG Face network with systems that use Euclidean or cosine distance between gallery templates and extracted deep features.
We demonstrate that iterative LOTS is very effective and show that systems utilizing deep features are easier to attack than the end-to-end network.

\end{abstract}

%%%%%%%%% BODY TEXT

\input{introduction}
\input{relatedwork}
\input{approach}
\input{experiments}

\input{conclusion}

%\ifbtasfinal
\section*{Acknowledgments}

This research is based upon work funded in part by NSF IIS-1320956 and in part by the Office of the Director of National Intelligence (ODNI), Intelligence Advanced Research Projects Activity (IARPA), via IARPA R\&D Contract No. 2014-14071600012. The views and conclusions contained herein are those of the authors and should not be interpreted as necessarily representing the official policies or endorsements, either expressed or implied, of the ODNI, IARPA, or the U.S. Government. The U.S. Government is authorized to reproduce and distribute reprints for Governmental purposes notwithstanding any copyright annotation thereon.
%\fi

{\small
\bibliographystyle{ieee}
\bibliography{ijcb2017}
}

\end{document}

%% file: introduction.tex
\section{Introduction}

\begin{figure}[t]
\vspace*{-0.1ex}
  \centering

  \ic[.15]{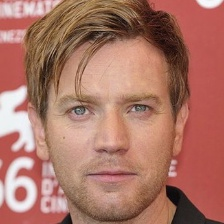} \thinspace \ic[.15]{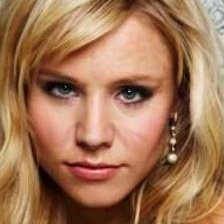} \thinspace \ic[.15]{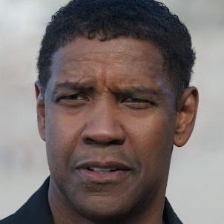}\\[0.75ex]

  \ic[.15]{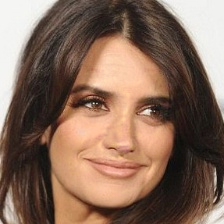} \thinspace \ic[.15]{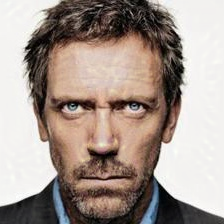} \thinspace \ic[.15]{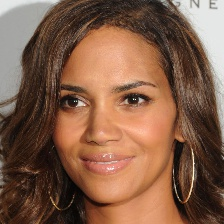}\\[0.75ex]

  \ic[.15]{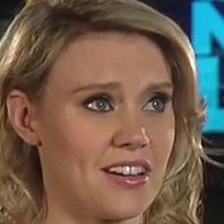} \thinspace \ic[.15]{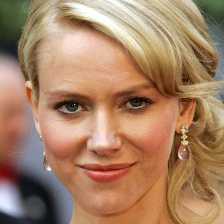} \thinspace \ic[.15]{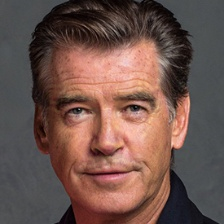}

  \cap{fig:teaser}{Who Are They?}{Although we might not be able to recognize the presented celebrities, we can still differentiate them from one another. In fact, eight of these images are manipulated in such a way that their VGG Face descriptors mimic Kate McKinnon's (bottom-left) and cause systems that apply Euclidean or cosine distance to classify each image incorrectly as her.}
\end{figure}

In the last few years, the most advanced deep neural networks (DNNs) have managed to reach or even surpass human level performance on a wide range of challenging machine learning tasks~\cite{parkhi2015deep,schroff2015facenet,szegedy2015going,he2015deep}, including face recognition.
DNNs trained to perform specific tasks are able to learn representations that generalize well to other datasets~\cite{taigman2014deepface,parkhi2015deep}, and the extracted generic descriptors can be utilized to tackle other diverse problems~\cite{sharif2014cnn, donahue2014decaf}.
For visual recognition tasks in biometrics -- e.g., facial attribute classification~\cite{liu2015deep} and face recognition~\cite{taigman2014deepface,sun2014deep,parkhi2015deep} -- features obtained from DNNs are widely used in the literature.

Although we are capable of designing and training DNNs that perform well, our understanding of these complex networks is still incomplete.
This was highlighted by the intriguing properties of machine learning models discovered by Szegedy \etal{szegedy2013intriguing}.
Namely, machine learning models -- including the state-of-the-art DNNs -- suffer from an unexpected instability as they misclassify adversarial examples formed by adding imperceptibly small perturbations to otherwise correctly recognized inputs.
Due to their excellent generalization capabilities, DNNs are expected to be robust to such small perturbations to their inputs, therefore the existence of adversarial examples challenges our understanding of DNNs and raises questions about the applications of such vulnerable learning models.

Considering the revealed adversarial instability of the end-to-end machine learning models, it is natural to ask whether systems utilizing extracted features from DNNs are also vulnerable to such perturbations.
In case they are susceptible to adversarial examples, are they more or less robust than end-to-end DNNs?
To be able to answer these questions, first we need to design a novel adversarial example generation technique that is capable of efficiently attacking those systems.

In this paper, we introduce the layerwise origin-target synthesis (LOTS) technique designed to perturb samples in such ways that their deep feature representations mimic any selected target activations.
We experimentally demonstrate the effectiveness of LOTS in terms of forming high quality adversarial examples.
We analyze and compare the robustness of the end-to-end VGG Face network~\cite{parkhi2015deep} to adversarial perturbations with other face recognition systems that utilize deep features extracted from the same network using Euclidean or cosine distance.
Our results show that LOTS is capable of successfully attacking each system, and that face recognition systems using the extracted deep features are less robust than the end-to-end network.

%% file: relatedwork.tex
\section{Related Work}

Automatic face recognition has a long history and many different approaches have been proposed in the literature \cite{tan2006face,jafri2009survey,serrano2010recent,gunther2016frice}.
While these traditional face recognition algorithms perform well on facial images with decent quality \cite{otoole2007face}, they are not able to handle pose variations \cite{gunther2016frice}.
Only the development of deep neural networks \cite{taigman2014deepface,sun2014learning,parkhi2015deep} has overcome this issue, and nowadays these methods are the quasi standard for face recognition in uncontrolled scenarios.
For example, the DNNs used by Chen \etal{chen2016unconstrained} provide the current state-of-the-art results on the IJB-A benchmark \cite{klare2015ijba}, which is outperformed by the (unpublished) DNN of Ranjan \etal{ranjan2017l2}.

In biometric recognition, training and evaluation can use different identities not just different images, which means that identities cannot be directly classified by an end-to-end network.
Instead, the last layer of the network is removed, and deep features extracted from the penultimate layer of the DNN are used as a representation of the face \cite{taigman2014deepface,parkhi2015deep}.
To form a more robust representation of an identity, deep features of several images are averaged \cite{chen2016unconstrained}.
Finally, the comparison of deep features is obtained via simple distance measures in the deep feature space such as Euclidean \cite{parkhi2015deep} or cosine distance \cite{chen2016unconstrained}.
Our experiments use deep features extracted with the publicly available VGG Face network \cite{parkhi2015deep}.

Since Szegedy \etal{szegedy2013intriguing} presented the problem posed by adversarial examples and introduced the first method capable of reliably finding such perturbations, various approaches were proposed in the literature.
Compared to the computationally expensive box-constrained optimization technique (L-BFGS) that Szegedy \etal{szegedy2013intriguing} used, a more lightweight, still effective technique was introduced by Goodfellow \etal{goodfellow2014explaining}.
Their fast gradient sign (FGS) method relies on using the sign of the gradient of loss with respect to the input, which needs to be calculated only once per adversarial example generation.
The authors demonstrated that using an enhanced objective function that implicitly incorporates FGS examples, the overall performance and the adversarial robustness of the trained models can be improved.
Later, Rozsa \etal{rozsa2016adversarial} showed that by not using the sign, the formalized fast gradient value (FGV) approach forms different adversarial samples than FGS and those yield a greater improvement when used for training.

The aforementioned two adversarial example generation techniques -- FGS and FGV -- rely on simply ascending the gradient of loss used for training the network. Namely, the formed perturbation causes misclassification by increasing the loss until the particular original class does not have the highest probability.
In their recent paper focusing on adversarial training, Kurakin \etal{kurakin2016adversarial} proposed extensions over the FGS method to be able to target a specific class or by calculating and applying gradients iteratively compared to a single one for conducting a line-search via FGS.

A few approaches that do not rely on using the gradient of training loss were also proposed by researchers.
Rozsa \etal{rozsa2016adversarial} introduced the hot/cold approach producing adversarial examples by both reducing the prediction probability of the original class of the input as well as increasing the probability of a specified target class.
To do so, the hot/cold approach defines a Euclidean loss with varying target classes on the pre-Softmax layer and uses its gradients as directions for forming adversarial perturbations.
This approach is capable of producing multiple adversarial examples per input, but still targets training classes, so cannot be directly applied to deep features.

Finally, the approach introduced by Sabour \etal{sabour2015adversarial} produces adversarial examples that not only cause misclassifications but also mimic the internal representations of the targeted inputs.
However, their technique relies on using the computationally expensive L-BFGS technique, which limits its application.

Since, in general, biometric systems operate on a dataset different than the end-to-end network was trained on, such systems cannot be attacked by end-to-end adversarial generation techniques.
Our novel LOTS method can be considered an extension of the hot/cold approach to deeper layers, and it also shows similarities to the technique of Sabour \etal{sabour2015adversarial} in terms of directly adjusting internal feature representations -- without relying on the L-BFGS algorithm.

%% file: approach.tex
\section{Approach}

This section describes the targeted face recognition systems, introduces our approach to form adversarial perturbations on those systems and, finally, presents the metric that we use for quantifying the quality of adversarial examples.

\subsection{Face Recognition Systems}
\label{sec:face_rec}

For our systems, we use the publicly available VGG Face dataset\footnote{\scriptsize\url{http://www.robots.ox.ac.uk/~vgg/data/vgg_face/}}~\cite{parkhi2015deep} which contains 2,604,175 images of 2,622 identities.
We chose this dataset because of its quality and size, and due to the fact that there is a publicly available end-to-end classification network, called the VGG Face network\footnote{\scriptsize\url{http://www.robots.ox.ac.uk/~vgg/software/vgg_face/}}~\cite{parkhi2015deep}, that was trained on this dataset.

The VGG Face network is intended to be used for extracting deep features, so-called VGG Face descriptors -- the authors successfully utilized the captured representations of the FC7 layer as face descriptors on the labeled face in the wild (LFW) dataset~\cite{parkhi2015deep} -- or for being fine-tuned on other datasets.
The network was trained on the VGG Face dataset, which was divided into three subsets containing 1,822,894, 520,835, and 260,446 images for training, validation, and test purposes, respectively.
Splitting the dataset happened proportionately with respect to images per identity: each identity in the dataset has $\leq$1000 images, 70$\%$, 20$\%$, and 10$\%$ of those are training, validation, or test images.

To be able to directly compare the robustness of the end-to-end VGG Face network with systems using the extracted VGG Face descriptors -- which are extracted at the FC7 layer before ReLU -- in latter systems we need to have the same identities as the VGG Face dataset has.
Therefore, we utilize the test set.
We form a gallery template for each identity by calculating the mean VGG Face descriptor of the first half of the test images ($\leq$50 per identity).
The VGG Face descriptors from the other half of the test images serve as probes, where each probe is compared to each gallery template, yielding 130,233 positive (same identity) and 341,340,693 negative (different identity) comparisons.

\begin{figure}[t]
\vspace*{0.5ex}
  %\centering
  \ic[.47]{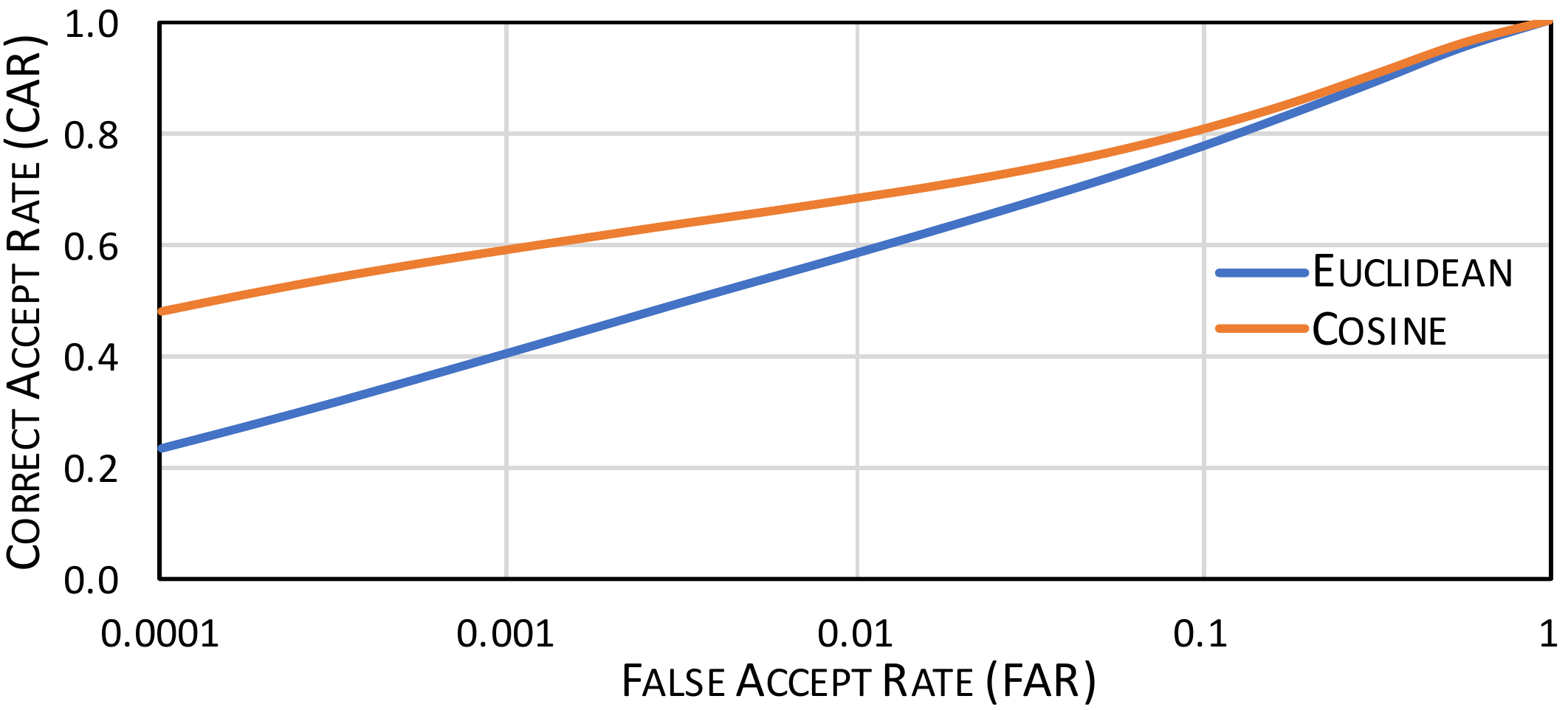}
  \cap{fig:roc}{ROC - Euclidean and Cosine Distances.}{Receiver operating characteristic (ROC) curves are shown for systems using Euclidean or cosine distance between gallery templates and VGG Face descriptors of probe images.}
\end{figure}

Using the positive and negative comparisons, we calculate Euclidean or cosine distance among them to compute ROC curves and, finally, we identify distance thresholds for attacking deep features in \sec{adv_scen}.
These ROC curves are displayed in \fig{roc}.
Since we would like to compare the adversarial robustness of the end-to-end network with systems having characteristics like real-world applications, we define them to have a low false accept rate (FAR) of 0.001 -- 1 successful zero-effort impostor attack out of 1000 attempts -- which translates into thresholds of 630.46 for Euclidean and 0.1544 for cosine distance.

\subsection{Attacking Deep Features with LOTS}

Let us consider a network $f$ with weights $w$ in a layered structure, i.e., having layers $y^{(l)}, l = \{1, \ldots,L\}$, with their respective weights $w^{(l)}$.
For a given input $x$, the output of the network can be formalized as:
\begin{equation}
  f(x) = y^{(L)}\left(y^{(L-1)}\left({\ldots\left({y^{(1)}(x)}\right)\ldots}\right)\right),
\end{equation}
while the internal representation (the deep feature) of the given input $x$ captured at layer $l$ is:
\begin{equation}
  f^{(l)}(x) = y^{(l)}\left(y^{(l-1)}\left(\ldots\left(y^{(1)}(x)\right)\ldots\right)\right).
\end{equation}

Our layerwise origin-target synthesis (LOTS) approach  adjusts the internal representation of an input $x_o$, the \textit{origin}, to get closer to the \emph{target} internal representation $t$.
In order to do so, we use a Euclidean loss defined on the internal representation $f^{(l)}(x_o)$ of the origin at layer $l$ and the target $t$, and apply its gradient with respect to the origin to manipulate the internal features of the origin, formally:
\begin{equation}
\label{lots}
%  \eta^{(l)}(x_o,x_t) = {{\nabla}_{x_o}} \left[ \frac{1}{2} \sum_i{{\left( f_i^{(l)}(x_t) - f_i^{(l)}(x_o) \right)}^2 } \right].
%  \eta^{(l)}(x_o,x_t) = {{\nabla}_{x_o}} \left( \frac{1}{2} \left\| f^{(l)}(x_t) - f^{(l)}(x_o) \right\| \right).
  \eta^{(l)}(x_o,t) = {{\nabla}_{x_o}} \left( \frac{1}{2} \left\| t - f^{(l)}(x_o) \right\|^2 \right).
\end{equation}

The target $t$ can be chosen without any constraints.
We can manipulate origin's features at layer $l$ to get closer to the feature representation of a specific targeted input $x_t$ using $t=f^{(l)}(x_t)$ or specify any arbitrary feature representation that the origin should mimic.

\begin{algorithm}[t]
\cap{alg:mimic_alg}{Deep Feature Mimicking Via LOTS.}{Iterative LOTS is a generic algorithm that perturbs origin $x_o$ in order to have deep features mimicking the specified target representation $t$. The function \texttt{mimicked} depends on the targeted system.}
\small
\begin{algorithmic}[1]
\Procedure{MIMIC}{$x_o,t$}\Comment{Origin $x_o$ mimics target $t$}
\State $x_p\gets x_o$
\State $x_p'\gets x_o$
\While{$ \text{not\,\,} \texttt{mimicked}(x_p, t)$}
\State $grad\gets \eta^{(l)}(x_p',t)$ \Comment{Eq.~\eqref{lots}}
\State $peak\gets \texttt{max}(\texttt{abs}(grad))$
\State $grad_s\gets grad\,/\,peak$ \Comment{Elementwise division}
\State $x_p'\gets \texttt{clip}(x_p' - grad_s)$
\State $x_p\gets \texttt{round}(x_p')$
\EndWhile
\State \textbf{return} $x_p$ \Comment{Image with features mimicking $t$}
\EndProcedure
\end{algorithmic}
\end{algorithm}

We can use the direction defined by the gradient of the Euclidean loss and form adversarial perturbations using a line-search -- similar to the fast gradient sign (FGS) method~\cite{goodfellow2014explaining} or the hot/cold approach~\cite{rozsa2016adversarial}.
Compared to those previous techniques, LOTS has the potential to form dramatically greater quantities of diverse perturbations for each input due to the billions of possible targets and the number of layers it can be applied on.
To form higher quality adversarial examples -- with less perceptible perturbations -- we can use iterative LOTS  as detailed in \alg{mimic_alg}.
This ``step-and-adjust'' algorithm perturbs $x_p$ initialized with the origin $x_o$ to get closer to the target $t$ step by step until the origin mimics the target, i.e., the Euclidean or cosine distance between $f^{(l)}(x_o)$ and $t$ is smaller than a predefined threshold value, or $x_p$ is classified by the end-to-end network as desired.
The perturbed image $x_p$ mimicking the target $t$ has discrete pixel values in $\left[0,255\right]$, however, while taking steps towards the target, the algorithm is designed to temporarily utilize non-discrete pixel values within $x_p'$ in order to obtain better adversarial quality.
Finally, note that we apply a scaled gradient (line 7 in \alg{mimic_alg}) with L$_\infty=1$ to move faster towards the specified target.

\subsection{Quantifying Adversarial Quality}

In order to analyze and compare the various attacks described in \sec{adv_scen}, we need to assess the quality of adversarial images that iterative LOTS can generate.
While L$_p$ norms are commonly used to quantify perturbations, some researchers~\cite{sabour2015adversarial,rozsa2016adversarial} concluded that those measures are not matched well to human perception.
To address the problem, Rozsa et al.~\cite{rozsa2016adversarial} proposed the psychometric called the perceptual adversarial similarity score (PASS) to better measure the quality of adversarial images.
While L$_2$ and L$_\infty$ norms focus strictly on the perturbation -- regardless of how visible it is on the distorted image -- PASS is designed to better quantify the distinguishability or similarity of the original image $x_o$ and the perturbed image $x_p$ with respect to human perception.

The calculation of PASS takes two steps: alignment by maximizing the enhanced correlation coefficient (ECC)~\cite{evangelidis2008parametric} of the image pair with homography transform $\Psi(x_p, x_o)$, followed by quantifying the similarity between the aligned original and perturbed images using the structural similarity (SSIM) index~\cite{wang2004image}.
By design, the alignment via ECC takes place before SSIM calculation as small translations or rotations can remain imperceptible to the human eye, thus, PASS eliminates those before determining the structural similarity of the image pair.
Consequently, PASS can be formalized as:
\begin{equation}
\label{pass}
\textup{PASS}\left( x_p, x_o \right) = \textup{SSIM}\left( \Psi \left( x_p,x_o \right),\, x_o \right),
\end{equation}
where $\textup{PASS}(x_p, x_o)=1$ indicates perfect similarity.

As the structural similarity via SSIM can be calculated only on grayscale images, we align the converted grayscale images using OpenCV's ECC with termination criteria of 100 iterations or $\epsilon=0.01$, then we calculate the structural similarity of the aligned images using SSIM.\footnote{Python implementation of SSIM by Antoine Vacavant: \scriptsize\\\url{http://isit.u-clermont1.fr/~anvacava/codes/ssim.py}}

%SSIM\footnote{Python implementation of SSIM by Antoine Vacavant: \url{http://isit.u-clermont1.fr/~anvacava/codes/ssim.py}}
%ECC\footnote{Image Alignment (ECC) in OpenCV by Satya Mallick: \url{http://www.learnopencv.com/image-alignment-ecc-in-opencv-c-python/}}

%% file: experiments.tex
\section{Experiments}

The primary goal of this paper is to answer the question whether systems relying on extracted deep features of DNNs are vulnerable to adversarial perturbations, and if they are, how their adversarial robustness compares to end-to-end classification networks'.
To be able to conduct a fair comparison, we need to design our experiments carefully.

\begin{figure*}[t]
  \vspace*{-1.5ex}
  \centering

  \subfloat[][Daniel Craig]{\ic[.155]{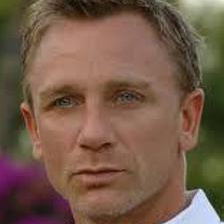}\label{adversaries:a}} \hspace{2pt}
  \subfloat[][Hugh Laurie]{\ic[.155]{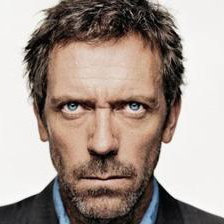}\label{adversaries:b}} \hspace{2pt}
  \subfloat[][Idris Elba]{\ic[.155]{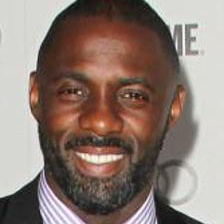}\label{adversaries:c}} \hspace{2pt}
  \subfloat[][Kate Beckinsale]{\ic[.155]{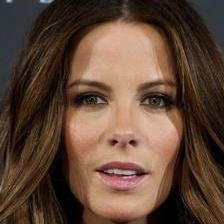}\label{adversaries:d}} \hspace{2pt}
  \subfloat[][Kristen Bell]{\ic[.155]{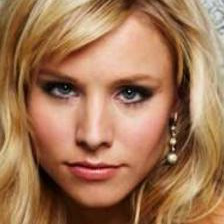}\label{adversaries:e}} \hspace{2pt}
  \subfloat[][Thandie Newton]{\ic[.155]{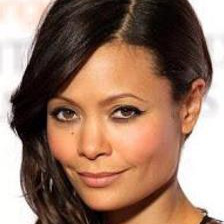}\label{adversaries:f}} \\

  \subfloat[][Denzel Washington]{\ic[.155]{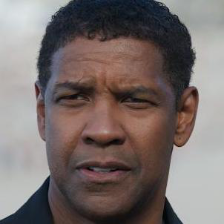}\label{adversaries:g}} \hspace{2pt}
  \subfloat[][Ewan McGregor]{\ic[.155]{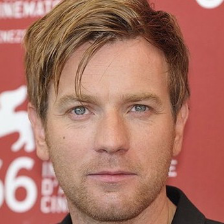}\label{adversaries:h}} \hspace{2pt}
  \subfloat[][Halle Berry]{\ic[.155]{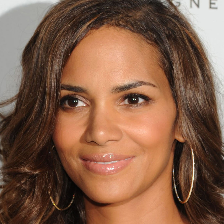}\label{adversaries:i}} \hspace{2pt}
  \subfloat[][Naomi Watts]{\ic[.155]{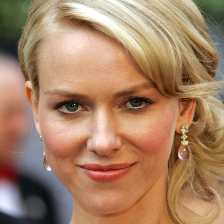}\label{adversaries:j}} \hspace{2pt}
  \subfloat[][Penelope Cruz]{\ic[.155]{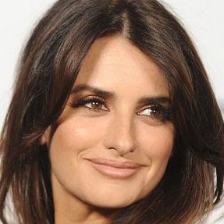}\label{adversaries:k}} \hspace{2pt}
  \subfloat[][Pierce Brosnan]{\ic[.155]{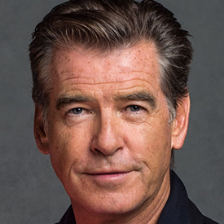}\label{adversaries:l}}

  \cap{fig:adversaries}{Adversaries - Internal and External.}{These are the adversaries that we use throughout our experiments. The internal adversaries shown in the top row are images from the VGG Face dataset that are correctly classified by each of our systems. The external adversaries displayed in the bottom are not contained in the VGG Face dataset.}
\end{figure*}
\vspace*{0.25ex}

\subsection{Adversaries and Attack Scenarios}
\label{sec:adv_scen}

For analyzing the capabilities of LOTS and studying the adversarial robustness of various face recognition systems, we use a dozen adversaries -- 6 identities hand-picked from the VGG Face dataset, along with 6 manually chosen external identities not contained in the VGG Face dataset.
We manually selected them in order to obtain a diverse set of adversaries as shown in \fig{adversaries}.
As each adversary is represented by a single image, internal adversaries need to be chosen carefully.
Therefore, from the validation set of the VGG Face dataset we selected an image for each internal adversary that is correctly classified by the end-to-end VGG Face network, and by both systems using Euclidean or cosine distance between the gallery templates and the extracted VGG Face descriptors of probe images, cf.~\sec{face_rec}.
Images representing the external adversaries were hand-picked and manually cropped.

With having both internal and external adversaries, our goal is to analyze whether VGG Face descriptors generalize well to novel identities or if they are more specific to the VGG Face dataset in terms of better representing those identities present in the dataset.
In the latter case, attacks conducted with external adversaries would outperform attacks by internal adversaries.

We conduct four sets of experiments utilizing the iterative layerwise origin-target synthesis (LOTS) approach, as detailed by \alg{mimic_alg}. We perturb images of adversaries such that they mimic the VGG Face dataset identities yielding misclassifications on different face recognition systems.

\emph{First}, on the end-to-end VGG Face network, we use LOTS on representations extracted from the Softmax layer.
While origins are external and internal adversaries, we specify the targeted identity by using a particular one-hot vector on the Softmax layer as target $t$.
This can be considered a traditional or more conventional approach for forming adversarial perturbations on end-to-end classification networks.
In fact, this utilization of LOTS can be interpreted as a slightly adjusted, iterative variant of the hot/cold approach introduced by Rozsa \etal{rozsa2016adversarial}.
\emph{Second}, we aim to generate adversarial perturbations on the end-to-end network using iterative LOTS on VGG Face descriptors of adversaries to mimic gallery templates that we formed by using the mean face descriptors of VGG identities (cf.~\sec{face_rec}).
This scenario can be viewed as attackers computing mean face descriptors from several images of targeted identities -- e.g., taken from the Internet -- and using them as target $t$.
We conduct these two experiments to assess the effectiveness of LOTS on the end-to-end VGG Face network.
The results also allow comparison of the more traditional approach of manipulating representations of the Softmax layer with the novel approach of mimicking VGG Face descriptors.

\begin{table*}[t]
\cap{tab:results}{Adversarial Example Generation Via Iterative LOTS.}{These results are obtained using iterative LOTS with the listed internal and external adversaries. With each adversary, we attacked every possible subject by mimicking their gallery templates to cause misclassifications on the end-to-end VGG Face network (End-To-End FD), and on systems using Euclidean or cosine distance between gallery templates and the extracted VGG Face descriptors. Furthermore, we attacked each identity on the end-to-end network by manipulating their representations at the Softmax layer (End-To-End SM) targeting the appropriate one-hot vector. We list the mean and standard-deviation of PASS, followed by the percentage of successful attacks, i.e., when the perturbed images were classified as the target.}
\vspace*{-1ex}
\centering
\scriptsize
\setlength{\tabcolsep}{3pt}
\resizebox{.98\textwidth}{!}{%
\begin{tabular}{c|l|c|c|c|c}
  \toprule
  \multicolumn{2}{c|}{\textsc{Adversary}} & \textsc{End-To-End SM} & \textsc{End-To-End FD} & \textsc{Euclidean Distance} & \textsc{Cosine Distance} \\
  \midrule
  \multirow{6}{*}{\rotatebox[origin=c]{90}{{\textsc{Internal}}}} &
  Daniel Craig 		& $0.9833\pm0.0076\,\,(98.44\%)$ & $0.9846\pm0.0075\,\,(98.32\%)$ & $0.9873\pm0.0083\,\,(100.00\%)$ & $0.9900\pm0.0055\,\,(100.00\%)$\\
  & Hugh Laurie 		& $0.9606\pm0.0186\,\,(98.44\%)$ & $0.9697\pm0.0123\,\,(98.32\%)$ & $0.9805\pm0.0116\,\,(100.00\%)$ & $0.9850\pm0.0081\,\,(100.00\%)$\\
  & Idris Elba 		& $0.9643\pm0.0206\,\,(98.36\%)$ & $0.9686\pm0.0147\,\,(98.32\%)$ & $0.9844\pm0.0130\,\,(100.00\%)$ & $0.9894\pm0.0075\,\,(100.00\%)$\\
  & Kate Beckinsale 	& $0.9804\pm0.0113\,\,(98.44\%)$ & $0.9840\pm0.0107\,\,(98.32\%)$ & $0.9900\pm0.0066\,\,(100.00\%)$ & $0.9921\pm0.0049\,\,(100.00\%)$\\
  & Kristen Bell 	& $0.9704\pm0.0156\,\,(98.44\%)$ & $0.9821\pm0.0115\,\,(98.32\%)$ & $0.9883\pm0.0062\,\,(100.00\%)$ & $0.9905\pm0.0049\,\,(100.00\%)$\\
  & Thandie Newton 	& $0.9792\pm0.0099\,\,(98.44\%)$ & $0.9849\pm0.0085\,\,(98.28\%)$ & $0.9881\pm0.0068\,\,(100.00\%)$ & $0.9904\pm0.0055\,\,(100.00\%)$\\
  \midrule
  \multirow{6}{*}{\rotatebox[origin=c]{90}{{\textsc{External}}}} &
  Denzel Washington	& $0.9866\pm0.0107\,\,(98.44\%)$ & $0.9839\pm0.0093\,\,(98.32\%)$ & $0.9869\pm0.0095\,\,(100.00\%)$ & $0.9900\pm0.0060\,\,(100.00\%)$\\
  & Ewan McGregor 	& $0.9925\pm0.0063\,\,(98.44\%)$ & $0.9936\pm0.0042\,\,(98.32\%)$ & $0.9944\pm0.0041\,\,(100.00\%)$ & $0.9957\pm0.0027\,\,(100.00\%)$\\
  & Halle Berry 		& $0.9913\pm0.0066\,\,(98.44\%)$ & $0.9918\pm0.0057\,\,(98.32\%)$ & $0.9943\pm0.0040\,\,(100.00\%)$ & $0.9955\pm0.0027\,\,(100.00\%)$\\
  & Naomi Watts 		& $0.9721\pm0.0150\,\,(98.44\%)$ & $0.9823\pm0.0100\,\,(98.32\%)$ & $0.9869\pm0.0071\,\,(100.00\%)$ & $0.9891\pm0.0065\,\,(100.00\%)$\\
  & Penelope Cruz 	& $0.9745\pm0.0151\,\,(98.44\%)$ & $0.9867\pm0.0084\,\,(98.32\%)$ & $0.9906\pm0.0052\,\,(100.00\%)$ & $0.9923\pm0.0048\,\,(100.00\%)$\\
  & Pierce Brosnan 	& $0.9835\pm0.0106\,\,(98.44\%)$ & $0.9808\pm0.0095\,\,(98.28\%)$ & $0.9877\pm0.0085\,\,(100.00\%)$ & $0.9907\pm0.0056\,\,(100.00\%)$\\
  \bottomrule
\end{tabular}%
}
\vspace*{1ex}
\end{table*}

\emph{Third} and \emph{fourth}, we conduct experiments to generate adversarial examples on face recognition systems that use Euclidean or cosine distance between the gallery templates and the extracted VGG Face descriptors of probe images.
Using iterative LOTS, our goal is to get face descriptors of adversaries closer to templates than Euclidean or cosine distance thresholds of systems having $\mathrm{FAR}=0.001$, as defined in \sec{face_rec}.

Throughout our experiments, we attempt to target every possible identity of the VGG Face dataset with each adversary.
For internal adversaries, this yields 2,621 subjects, while external adversaries can aim at impersonating all 2,622 identities.
To limit the computational costs, we constrain iterative LOTS to 500 steps.
In case the algorithm exceeds the limit, the particular attempt is considered a failure.
As we will see, this constraint has little effect on our experiments.
Furthermore, based on our experience, iterative LOTS taking more than 500 steps produces perturbations that are highly visible, in other words, those examples are not adversarial at all.

\subsection{Results}

The results obtained by conducting the four sets of experiments using the selected adversaries are presented in \tab{results}.
Comparing the collected metrics on the two types of attacks on the end-to-end VGG Face network, we can conclude that, in general, iterative LOTS operating on VGG Face descriptors produces examples with better adversarial quality than the traditional attack working on the Softmax layer.
Considering all internal and external adversaries, there is only one exception: for external adversary Denzel Washington, the formed examples using Softmax features contain less perceptible perturbations in average, as indicated by the higher PASS.
Furthermore, we can note with respect to the attacks on the end-to-end face recognition network that there is a small proportion of targeted identities for each adversary -- varying between 41 and 45 -- where iterative LOTS limited to 500 steps failed.
By analyzing these unsuccessful attempts, interestingly, we find that using the diverse set of adversaries our algorithm failed to form perturbations more or less for the same targeted identities.
We conjecture that those subjects are hard to reach via iterative LOTS because they are simply more difficult to be recognized by the end-to-end network  -- Doddington \etal{doddington1998sheep} dubbed them as ``goats.''
Consequently, those directions provided by the calculated gradients via iterative LOTS simply cannot find a way to those identities.

We can see in \tab{results} that iterative LOTS performs better on face recognition systems that use Euclidean or cosine distance on extracted VGG Face descriptors with $\mathrm{FAR}=0.001$ than it does on the end-to-end system.
The better performance is highlighted by both the higher percentage of successful attacks and the overall adversarial quality shown by generally higher PASS.
Statistical testing -- two-sided heteroscedastic t-tests with Bonferroni correction -- show very significant ($p<0.00001$) difference between each pair of the four attack scenarios. While the differences in PASS may seem to be small, the large sample size of over 2,600 identities results in the strong rejection of the hypothesis that the four attacks provide similar results.
For the methods shown in \tab{results}, the quality of the generated adversarial images statistically significantly increases from left to right, supporting the conclusion: {\em Systems utilizing deep features are easier to attack and admit less perceptible perturbations than the end-to-end network}.

While iterative LOTS forms perturbations for adversaries to reach nearly all targeted identities, the manipulated images also maintain high adversarial quality.
To demonstrate the effectiveness of iterative LOTS in terms of reaching the targeted subjects via small distortions, in \fig{advs} we show examples with VGG Face descriptors that are closer to gallery templates than Euclidean or cosine thresholds.
Most of these examples are indeed adversarial images as those perturbations are imperceptible.
The displayed examples are external adversaries targeting identities of the selected images used as internal adversaries.
We show these particular examples simply because we already introduced those identities by displaying an image for each internal adversary in \fig{adversaries}, thus we can associate a face to those subjects.

\begin{figure*}[t]
%\vspace{0.5ex}
  \centering

    \ic[.155]{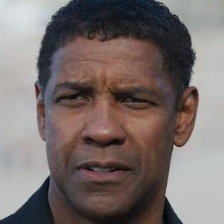} \thinspace
    \ic[.155]{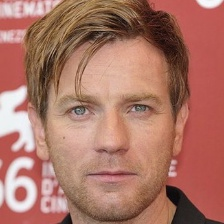} \thinspace
    \ic[.155]{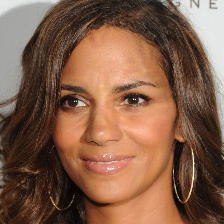} \thinspace
    \ic[.155]{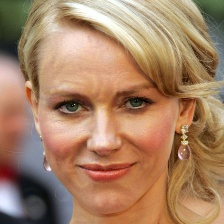} \thinspace
    \ic[.155]{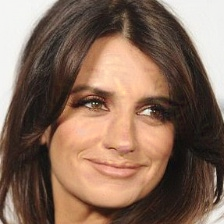} \thinspace
    \ic[.155]{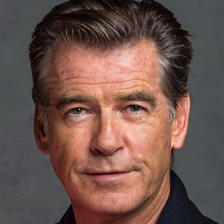} \\[0.75ex]

    \ic[.155]{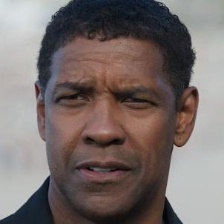} \thinspace
    \ic[.155]{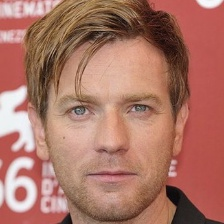} \thinspace
    \ic[.155]{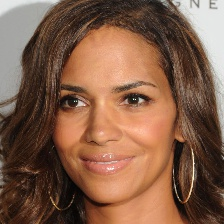} \thinspace
    \ic[.155]{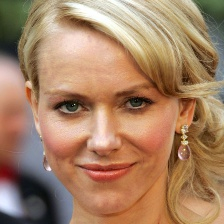} \thinspace
    \ic[.155]{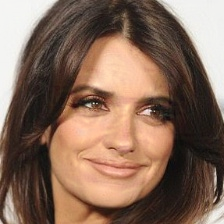} \thinspace
    \ic[.155]{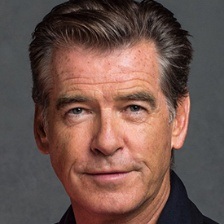} \\[0.75ex]

    \ic[.155]{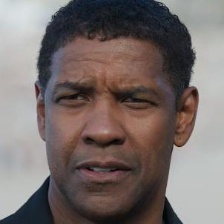} \thinspace
    \ic[.155]{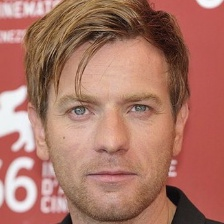} \thinspace
    \ic[.155]{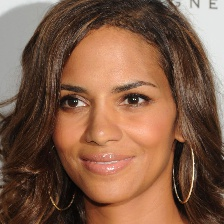} \thinspace
    \ic[.155]{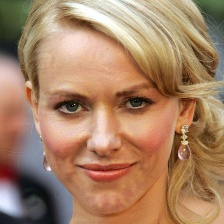} \thinspace
    \ic[.155]{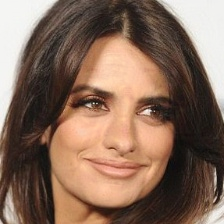} \thinspace
    \ic[.155]{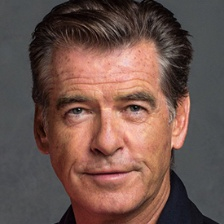} \\[0.75ex]

    \ic[.155]{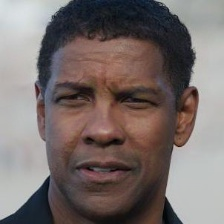} \thinspace
    \ic[.155]{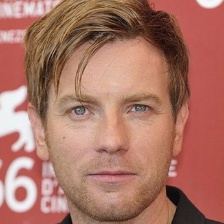} \thinspace
    \ic[.155]{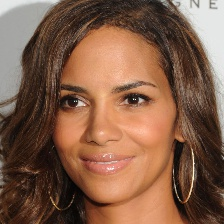} \thinspace
    \ic[.155]{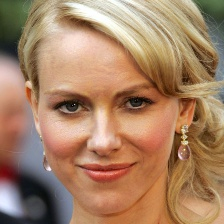} \thinspace
    \ic[.155]{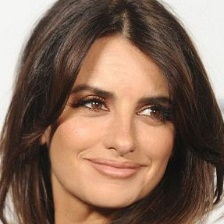} \thinspace
    \ic[.155]{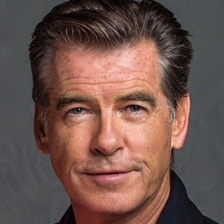} \\[0.75ex]

    \ic[.155]{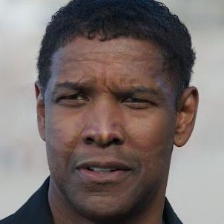} \thinspace
    \ic[.155]{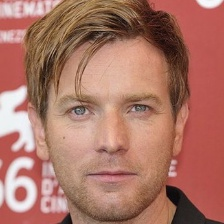} \thinspace
    \ic[.155]{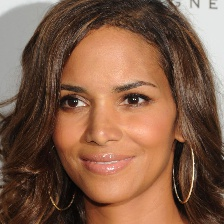} \thinspace
    \ic[.155]{Naomi_Watts_n00001407_9978} \thinspace
    \ic[.155]{Penelope_Cruz_n00001407_9963} \thinspace
    \ic[.155]{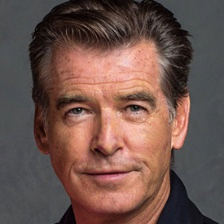} \\[0.75ex]

    \ic[.155]{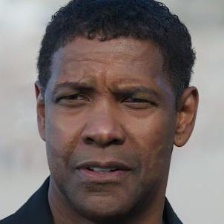} \thinspace
    \ic[.155]{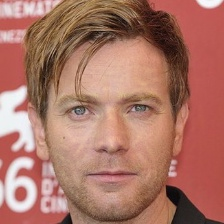} \thinspace
    \ic[.155]{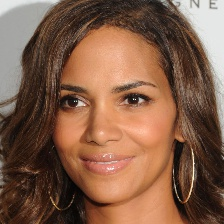} \thinspace
    \ic[.155]{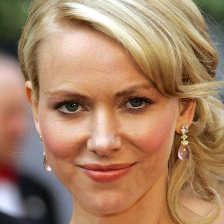} \thinspace
    \ic[.155]{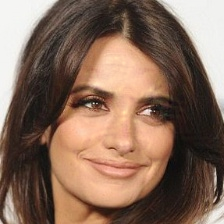} \thinspace
    \ic[.155]{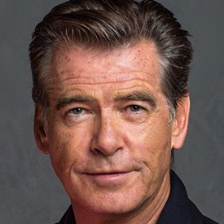}

  \cap{fig:advs}{Iterative LOTS on VGG Face Descriptors with External Adversaries.}{These perturbed images of external adversaries mimic targeted gallery templates. The VGG Face descriptors of all examples are incorrectly verified to match the gallery templates with Euclidean or cosine distances below $\mathrm{FAR}=0.001$ thresholds: from top row to bottom, images match gallery templates of Daniel Craig, Hugh Laurie, Idris Elba, Kate Beckinsale, Kristen Bell, and Thandie Newton (cf.~\fig{adversaries}).}
\end{figure*}

As indicated by the collected metrics, the formed perturbations that we obtained on the system using cosine distance yield even slightly better adversarial quality than collected on the Euclidean system.
This means that the system with the higher recognition accuracy (cf.~\fig{roc}) is also easier to attack.
To highlight the differences among systems with respect to adversarial vulnerability, we visualize perturbations for some distorted examples to show what it takes to cause misclassifications.
These can be seen in \fig{adv_examples}, where we display perturbed examples causing misclassifications on the three systems by manipulating VGG Face descriptors via iterative LOTS.
To be able to directly compare the various distortions, we show PASS as well as L$_2$ and L$_\infty$ norms of perturbations in sub-captions.

Finally, we can observe that the collected metrics on the produced examples generated via iterative LOTS vary among adversaries.
While we cannot see a trend differentiating internal and external adversaries, the distorted examples produced using the various adversaries have significantly different adversarial qualities.
We believe this is normal -- a face close to the average is naturally closer to others, contrarily, a very characteristic face is farther away and, thus, needs stronger perturbations to be turned to others.
For example, as the internal adversary of Hugh Laurie has a unique and very characteristic face among adversaries, it is not surprising that the distorted images of that adversary have one of the worst overall adversarial qualities.
On the other hand, we have two external adversaries -- Halle Berry and Ewan McGregor -- that can be easily turned to other subjects with smaller, less perceptible perturbations relative to other adversaries.
Doddington \etal{doddington1998sheep} referred to such identities as ``wolves.''

\begin{figure*}[t]
\vspace{-1.5ex}
  \centering

  \subfloat[][\label{adv_examples:a}\centering \textsc{End-To-End Network} \par PASS{\,=\,}0.9940\,\,\,L$_2${\,=\,}572.67\,\,\,L$_{\infty}${\,=\,}22]{\ic[.155]{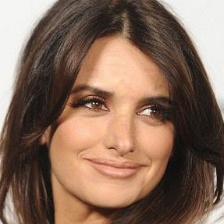}\,\ic[.155]{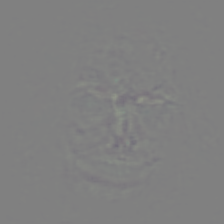}} \hspace{5pt}
  \subfloat[][\label{adv_examples:b}\centering \textsc{Euclidean Distance} \par PASS{\,=\,}0.9963\,\,\,L$_2${\,=\,}429.98\,\,\,L$_{\infty}${\,=\,}16]{\ic[.155]{Penelope_Cruz_n00001407_9963}\,\ic[.155]{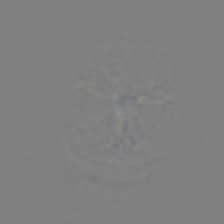}} \hspace{5pt}
  \subfloat[][\label{adv_examples:c}\centering \textsc{Cosine Distance} \par PASS{\,=\,}0.9970\,\,\,L$_2${\,=\,}386.40\,\,\,L$_{\infty}${\,=\,}14]{\ic[.155]{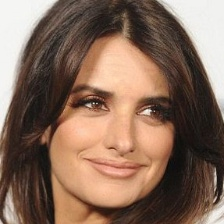}\,\ic[.155]{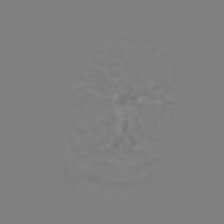}} \\

  \subfloat[][\label{adv_examples:d}\centering \textsc{End-To-End Network} \par PASS{\,=\,}0.9703\,\,\,L$_2${\,=\,}1526.07\,\,\,L$_{\infty}${\,=\,}64]{\ic[.155]{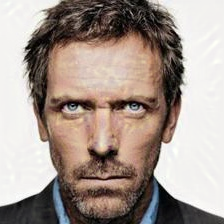}\,\ic[.155]{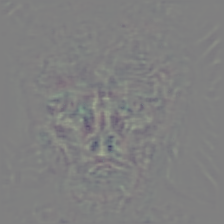}} \hspace{5pt}
  \subfloat[][\label{adv_examples:e}\centering \textsc{Euclidean Distance} \par PASS{\,=\,}0.9866\,\,\,L$_2${\,=\,}947.47\,\,\,L$_{\infty}${\,=\,}41]{\ic[.155]{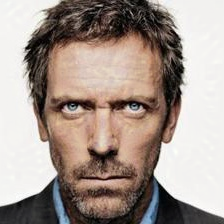}\,\ic[.155]{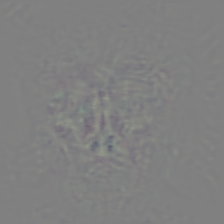}} \hspace{5pt}
  \subfloat[][\label{adv_examples:f}\centering \textsc{Cosine Distance} \par PASS{\,=\,}0.9879\,\,\,L$_2${\,=\,}899.18\,\,\,L$_{\infty}${\,=\,}38]{\ic[.155]{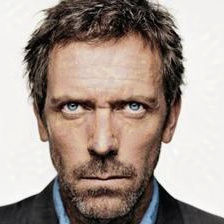}\,\ic[.155]{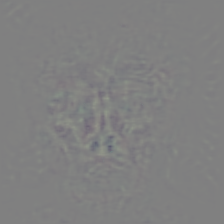}} \\

%  \subfloat[][\label{adv_examples:g}\centering \textsc{End-To-End Network} \par PASS{\,=\,}0.9963\,\,\,L$_2${\,=\,}465.88\,\,\,L$_{\infty}${\,=\,}30]{\ic[.15]{Naomi_Watts_n00001407_9963}\,\ic[.15]{Naomi_Watts_n00001407_9963_noise}} \hspace{5pt}
%  \subfloat[][\label{adv_examples:h}\centering \textsc{Euclidean Distance} \par PASS{\,=\,}0.9979\,\,\,L$_2${\,=\,}342.03\,\,\,L$_{\infty}${\,=\,}22]{\ic[.15]{Naomi_Watts_n00001407_9978}\,\ic[.15]{Naomi_Watts_n00001407_9978_noise}} \hspace{5pt}
%  \subfloat[][\label{adv_examples:i}\centering \textsc{Cosine Distance} \par PASS{\,=\,}0.9984\,\,\,L$_2${\,=\,}286.56\,\,\,L$_{\infty}${\,=\,}18]{\ic[.15]{Naomi_Watts_n00001407_9984}\,\ic[.15]{Naomi_Watts_n00001407_9984_noise}}

  \cap{fig:adv_examples}{Adversarial Examples Via Iterative LOTS on VGG Face Descriptors Targeting Kristen Bell.}{This figure shows adversarial examples paired with their corresponding perturbations that yield incorrect classifications on the end-to-end VGG Face network, and on systems using Euclidean or cosine distance between the extracted VGG Face descriptors and the gallery template of Kristen Bell. The sub-captions show the targeted system, the PASS between the origin and the perturbed image, and the L$_2$ and L$_\infty$ norms of the perturbation.}
\vspace{1.5ex}
\end{figure*}

%% file: conclusion.tex
\section{Conclusion}

Since researchers mainly focus on adversarial example generation techniques and, in general, adversarial robustness on end-en-end classification networks, the primary goal of this paper was to extend research to systems that utilize deep features extracted from deep neural networks (DNNs), which is common in biometrics.
In this paper, we have introduced our novel layerwise origin-target synthesis (LOTS) algorithm.
LOTS is generic and can be efficiently used iteratively to form adversarial examples both on end-to-end classification networks and on systems that use extracted deep features of DNNs.

We have experimentally demonstrated the capabilities of iterative LOTS by generating high quality adversarial examples on different systems.
We have conducted large-scale experiments to compare the adversarial robustness of three face recognition approaches using a dozen adversaries targeting all possible identities.
We have generated adversarial examples on the end-to-end VGG Face network via iterative LOTS working on the extracted VGG Face descriptors, and, more traditionally, on features of the Softmax layer.
Furthermore, using iterative LOTS, we have formed adversarial perturbations on systems that use VGG Face descriptors with Euclidean or cosine distance that are closer to the targeted gallery templates than the $\mathrm{FAR}=0.001$ thresholds.

To assess the robustness of the targeted systems, we have quantified the quality of the produced adversarial examples using the perceptual adversarial similarity score (PASS), and we have measured the percentage of successful attempts where the perturbed images are classified as the targeted identities.
A less vulnerable system allows adversaries to impersonate fewer of their targeted identities and/or requires adversaries to form stronger, thus more visible perturbations in order to achieve the targeted misclassifications.
Based on the collected metrics, we have concluded that the end-to-end system is more robust to adversarial perturbations formed by iterative LOTS, and the system utilizing cosine distance is the most vulnerable among all.
While adversaries could not reach all their targeted identities on the end-to-end VGG Face network, they could achieve that on the other systems utilizing the extracted face descriptors -- along with better adversarial qualities.
Unfortunately, the system most vulnerable to iterative LOTS is preferred in biometrics due to the fact that, in general, cosine distance provides better performing systems than those that utilize Euclidean distance.

Finally, although we have performed our experiments only using VGG Face descriptors to form adversarial examples, we assume that our results will be portable to other network architectures.
We were only targeting ``raw'' deep features, while deep features are often processed by triplet-loss embedding \cite{parkhi2015deep,sankaranarayanan2016triplet} before the applicable distances are calculated.
As these projections are external to the DNN, and the triplet-loss projection matrix from Parkhi \etal{parkhi2015deep} is not available, we cannot attack deep features after triplet-loss embedding.
We conjecture that iterative LOTS is capable of forming examples causing incorrect recognition on systems applying triplet-loss embedding or lower FAR thresholds, the only question is whether the produced examples would be adversarial in terms of human perception.
In future work, we will consider attacking such face recognition systems.